\documentclass{article}

\usepackage{arxiv}

\usepackage[utf8]{inputenc} 
\usepackage[T1]{fontenc}    
\usepackage{hyperref}       
\usepackage{url}            
\usepackage{booktabs}       
\usepackage{amsfonts}       
\usepackage{nicefrac}       
\usepackage{microtype}      
\usepackage{lipsum}
\usepackage{graphicx}
\usepackage{amsmath}
\usepackage{algpseudocode}
\usepackage{algorithm}
\usepackage{pbox}
\usepackage{mathtools}
\usepackage{adjustbox}
\usepackage[numbers]{natbib}
\usepackage[super]{nth}
\usepackage{url}
\newcommand{\floor}[1]{\left\lfloor #1 \right\rfloor}
\title{Simulation and Augmentation of Social Networks for Building Deep Learning Models}

\author{Akanda Wahid -Ul- Ashraf \\
  Department of Computing and Informatics\\
  Bournemouth University\\
  Fern Barrow, Poole BH12 5BB, UK\\
  \texttt{aashraf@bournemouth.ac.uk} \\
   \And
Marcin Budka \\
  Department of Computing and Informatics\\
  Bournemouth University\\
  Fern Barrow, Poole BH12 5BB, UK\\  
  \texttt{mbudka@bournemouth.ac.uk} \\
   \And
Katarzyna Musial \\
  Advanced Analytics Institute\\
  School of Software\\
  Faculty of Engineering and IT\\
  University of Technology Sydney\\
  Australia\\  
  \texttt{katarzyna.musial-gabrys@uts.edu.au}  }

\begin{document}
\maketitle

\begin{abstract}
Graph Convolutional Networks (GCNs), a neural network-based classification model on graphs, have been shown to outperform other state-of-the-art models. However, one major limitation of the GCN is that it assumes at a particular $l^{th}$ layer of the neural network model only the $l^{th}$ order neighbourhood nodes of a social network are influential. Furthermore, the GCN has been evaluated on citation and knowledge graphs, but not especially on friendship-based social graphs. Moreover, the drawback associated with the dependencies between layers and the order of node neighbourhood for the GCN can be more prevalent for friendship-based social graphs. The evaluation of the full potential of the GCN on friendship-based social network requires openly available datasets in larger quantities. However, most available social network datasets are not complete (i.e. represent a subset of the original networks, not the entire graph or do not include the entire set of node features). On top of that, the majority of the available social network datasets not only do not contain any features but also ground truth labels. To address the need for good quality synthetic social network data with ground truth labels and features we firstly provide a guideline on how to simulate dynamic social networks, with ground truth labels and features, both coupled with the topology of the network. Secondly, we introduce an open-source Python-based social network simulation library with GPU computation and multiprocessing~\footnote{https://github.com/AkandaAshraf/VirtualSoc}. In our social network simulation, we argue that the topology of the network is driven by a set of latent variables, termed as the social DNA (\emph{sDNA}). We consider the \emph{sDNA} as labels for the nodes, mimicking the real-world social network scenario. Finally, by evaluating on our simulated datasets, we propose four new variants of the GCN, mainly to overcome the limitation of dependency between the order of node-neighbourhood and a particular layer of the model. We then evaluate the performance of all the models and our results show that on 27 out of the 30 simulated datasets our proposed GCN variants outperform the original model.

\end{abstract}

\keywords{Graph Mining \and Node Classification \and Social Network Mining \and Deep Learning on Graphs \and Social Network Simulation}

\section{Introduction}
\label{intrduction}

One major limitation of a neural network-based learning systems is that they requires a large amount of data for training. This is one of the biggest differences between human intelligence and artificial non-general intelligence like an artificial neural network. Unlike a deep learning (i.e. deep neural network) model, a human can learn from a very limited number of examples, whereas a deep learning model requires to see a substantially larger number of samples to learn from. Thus, it is essential to have access to a large number of training data instances to unlock and evaluate the full potential of the neural network-based model. A straightforward technique to solve this problem of insufficiency of the real-world datasets for neural network-based learning systems is to simulate high quality real-world alike synthetic data and use it to train the model. Additionally, if not for training, simulated datasets are particularly useful to evaluate the models' performance, i.e. during the testing phase. In many cases, it is far more convenient to simulate test cases representing exceptional situations than collecting data for those situations in the real world. In fact, for some real-world scenarios, it might not even be possible to get a dataset describing some exceptional scenario due to the rarity of the event or ethical constraints.

It is however crucial to test the trained model in those exceptional scenarios because the cost of failure for those unlikely situations can be significantly higher than a regular situation. One such area where high quality simulated and augmented data is extensively being used are in the neural network-based learning systems for self-driving cars. Almost all the advanced autonomous vehicle technologies use simulated datasets. For example, Nvidia has developed the Nvidia Drive Constellation, a Virtual Reality Autonomous Vehicle Simulator~\cite{@nvidia}. Billions of miles have been driven in the simulated environment by Google's Waymo~\cite{Waymo} etc. Similar to the self-driving cars, in many other applications of deep learning, high quality simulated datasets are now in high demand. One such important application of deep learning, which is the focus of this study, is in the area of social networks, where graph specific deep learning models are ever-increasingly being developed and evaluated~\cite{goyal2018graph}. With the advancement of graph specific neural network-based models, the demand for such datasets is growing rapidly. Furthermore, it is becoming more and more difficult to have access to complete (i.e. inclusive of node attributes) datasets representing social networks mainly due to user privacy concerns that we discuss later in this section. 

Social network datasets are very complex in nature, thus, they can be difficult to simulate and there is a lack of comprehensive guidelines on how to simulate social network datasets with both the features and ground truth labels. 

As mentioned earlier, graph data mining has become a very important research area due to the recent advancement and popularity of social networks~\cite{wasserman1994social,newman2018networks}, especially the online ones. Advancements in graph-based predictive modelling or graph community detection algorithms require datasets with ground truth labels for evaluation purposes~\cite{sapountzi2018social}. However, majority of the available social network datasets do not contain labels. Moreover, real-world social network datasets contain high dimensional features (node attributes and features are used interchangeably in this paper)~\cite{pecli2018automatic} that represent information about both nodes and relationships. For example, a Facebook user generates variety of information such as posts he/she likes, photos, status updates, etc. Even in citation networks, there are features such as domain, authors' affiliations, documents with thousands of words, etc~\cite{popescul2003statistical}. In publicly available datasets, such features are rarely included. For a small number of datasets, these node attributes could be included but then usually the complete structure of the network is not; instead only its subset (mainly ego networks) are available~\cite{viswanath2009evolution,FBKONECT,snapnets}. 
This is due to the fact that during the anonymisation process of networked data, in most cases we need to get rid of majority of features as these could be used to identify individuals~\cite{townsend2016social}, potentially raising ethical concerns. De-identification of network datasets is particularly difficult because of the unique topological structure a network may have. In a 2011 Kaggle link prediction competition, the most successful team won by de-anonymising most of the network data~\cite{narayanan2011link}. On top of that, nowadays, even such graph datasets are becoming very difficult to obtain due to the aftermath of the notorious usage of real-world dataset from social networks for the purpose of political influence~\cite{hand2018aspects,cadwalladr2018revealed}. 


To ensure user's personal data is only used with explicit consent, governments and political unions are increasingly putting pressure on the technology companies~\cite{michelle_2018}. Additionally, new regulations such as the European General Data Protection Regulation (GDPR) on the usage of personal data, has already come into force in many countries such as the UK~\cite{bennett2018european}. Unquestionably, such regulations are essential to guarantee user privacy. However, due to those, getting hold of datasets from social media is becoming increasingly challenging. Maintaining the advancement of the research in social networks requires good quality real-world datasets. One solution is to supplement the real-world social network datasets with synthetic, good quality, real-world alike data. 

The demand for graph datasets is further on the rise, due to the advancement of graph-based machine learning, as traditional learning and data mining algorithms are being adopted for graph mining. Machine learning tasks for non-relational datasets only consider features and labels. However, graph datasets also contain edges between instances. These relationships have the ability to provide additional predictive power for a machine learning model. As a result, including these relations along with features in a predictive model is vital for prediction based on graph-structured datasets. To include relationships, one may capture these relational links between instances through graph embedding and then train any traditional machine learning model for the task of classification or regression~\cite{mikolov2013distributed,grover2016node2vec}. However, besides this indirect consideration of relations or links, there are developments in the area of graph mining which directly encode a relational component of a graph dataset into a deep artificial neural network, termed as Graph Convolutional Networks (GCNs)~\cite{kipf2016semi}. In this approach, the topology of a graph is directly translated into the layers of a deep learning model. In GCN, the features of the graph are multiplied with filters of a neural network in the spectral domain (i.e. graph Laplacian) of the graph, thus resulting in a direct convolutional operation. Apart from node classification which is one of the most researched problems in machine learning, an important research area in graph mining is link prediction. A difficulty encountered when analysing any link prediction technique is not being able to get enough, open, dynamic (time-dependent snapshots) social networks with features  and labels. Typically, a link prediction algorithm is tested based on its predictive power on a future snapshot of the network. A supervised link prediction algorithm should ideally utilise both the topology and available node attributes~\cite{lichtenwalter2010new,pecli2018automatic}. For example, \citet{scellato2011exploiting} found that including features such as places and other related user activity improves the accuracy of link prediction considerably. Most of the developments in link prediction have been based on a single snapshot of the network, although, incorporating evolution of the graph may result in better performance in link prediction as shown by~\citet{tylenda2009towards} and~\citet{xu2018supervised}.

GCN is a semi-supervised classification model shown to outperform other state-of-the-art graph classification approaches based on as little as 0.07\% of labelled nodes per class~\cite{kipf2016semi}. In the paper where GCN is introduced, datasets considered in the experiment were citation networks and knowledge graphs with explicitly defined class labels~\cite{kipf2016semi}. However, defining class labels for Facebook, Twitter, LinkedIn like social networks is not trivial. As discussed earlier, the difficulty is mainly associated with obtaining real-world graph datasets with labels and node attributes. One approach to evaluate such graph mining algorithms is by simulating graphs containing features. In this work, we propose that the preference of each node in a social network is the strongest, useful and meaningful candidate for label in a social graph. 


\section{Related work}

A straightforward way to simulate graphs is to generate them using well-established network models~\cite{wahid2018netsim}: (1)~Barab\'{a}si-Albert model for the scale-free network~\cite{barabasi1999emergence}, (2)~Watts-Strogatz small-world model for the small-world network~\cite{watts1998collective}, (3)~Erd\H{o}s-R\'{e}nyi model for the random graph network~\cite{solomonoff1951connectivity,erdos1959random,erdos1960evolution,erdHos1961strength}, (4)~Forest-fire Model model~\cite{leskovec2005graphs,drossel1992self}.

\textbf{Random-graph Model:} In the random graph network model, one creates a network with some properties of interest (specific degree distribution) and otherwise random. Although random graph model was first studied by~\citet{solomonoff1951connectivity} this model is mainly associated with Paul Erd\H{o}s and Alfr\'{e}d R\'{e}nyi~\cite{erdos1959random,erdHos1961strength}. 


\textbf{Scale-free Model:} The scale-free model shows power law node degree distribution $P(k) \sim k^{-\alpha}$ (where $k$ is the node degree and typically $2 < \alpha  < 3$) for a social network. This kind of distribution was first discussed by~\citet{price1976general}. Price, in turn, was inspired by Herbert Simon, who discusses power law in a variety of non-network economic data~\cite{simon1955class}.


\textbf{Small-world Model:} Transitivity measured by the network clustering coefficient despite being extensively studied, is still one of the least understood properties in network analysis according to~\citet{newmannetworks}. Another important property we observe in real networks is the small-world effect -- all nodes are connected with each other by relatively short paths. To model these two properties Watts and Strogatz introduced a small-world network model~\cite{watts1998collective}.


\textbf{Forest-fire Model:} In this model the new node, $i$, connects to another existing node $j$, and then again makes a connection with the adjacent node $j_1$ of the newly connected node $j_0$. The node $i$ then carries on making connections with a probability $p$ based on adjacent nodes~\cite{leskovec2005graphs,drossel1992self}. For example, in citation networks, an author finds a paper and cites it. He or she then cites more papers through that paper recursively~\cite{leskovec2005graphs}. In a social network, a friend $j$ may introduce someone $i$ with his/her mutual friend and then the friend circle grows for the person $i$~\cite{leskovec2005graphs}. The model is named as forest fire because it imitates self-organising behaviour of a forest fire~\cite{drossel1992self}. 


These quintessential network models are one of the most important contribution towards understanding and modelling complex networks. However, these mathematical models are solely driven by the topology of a network. For example, the Scale-free model considers the degree of a node and the Small-world model considers mutual friends. Neither features nor labels of nodes and/or connections are mimicked by those models. However, one can generate synthetic social networks with features is to find similarities/correlations between randomly assigned $n$ number of features and let those similarities define connections~\cite{papadimitriou2011predicting,symeonidis2010transitive,papadimitriou2012fast}. For obvious reasons, this na\"ive approach is not ideal due to several limitations. Firstly, correlations between feature vectors do not consider the network topology. Secondly, a common correlation metric would assume every person in a social network views and prefers a potential friend's features equally in a linear fashion. Finally, it is often not obvious what the node labels are, which is an issue we discuss in detail in Section~\ref{our approach}. 

However, there are some recent developments in agent and event-based social network modelling which are discussed below. 

\textbf{Agent-based modelling:} \citet{bruch2015agent} provide a guideline on the agent-based modelling of social networks. In the paper by~\citet{bruch2015agent}, it is argued that the interplay between the micro and macro level characteristics is complex, and the macro level characteristics are not emergent solely from the simple aggregation of micro level characteristics or low level entities such as social network users~\cite{granovetter1978threshold}. Instead, micro and macro level behaviour or characteristics form a feedback loop, resulting in a nonlinear interaction. From a social network point of view, the graph level and node level characteristics could be thought of as macro and micro level characteristics of the network respectively. 

However, to simulate the modelling approach specific to social networks, one should consider the well-studied graph properties such as preferential attachment, mutual friend preferences etc., and provide instructions on how to account for these properties in the simulation. In the research article of~\citet{granovetter1978threshold}, these social network properties are considered, but implicitly included in terms of the macro and micro level characteristics. In summary, this study by~\citet{granovetter1978threshold} provides a generic guideline to model social networks but a detailed and specific mathematical modelling instruction and analysis of the social network properties are not discussed. 


In another work by ~\citet{kavak2018big}, the authors have argued that modelling should be performed by explicitly using available real-world dataset. In their experiment, they have simulated human mobility model based on 826,021,868 twitter messages. Furthermore, they have uncovered the Geolocation of 92,296 users for the purpose of modelling. However, the purpose of our simulation is to produce synthetic good quality graph structured datasets when real-world data is not available, which is increasingly the case as discussed in Section~\ref{intrduction}.

\textbf{Event based modelling:} One recent interesting development in modelling dynamic event-based graph is the Cognition-driven Social Network (CogSNet) model~\cite{michalski2018social}. The CogSNet models social network-based on the human memory model. Authors argue that, similar to the human memory, a social event is strengthened by repeated exposure to a similar event and weakens by deprivation of that event. Although CogSNet proposes a new paradigm in social network modelling, it does not provide an explicit explanation modelling features within the dynamic event based graph. Providing open source social network datasets with labels, features, and graph or topological characteristics is the primary goal of this study. 

To address the issues discussed above, i.e., 1)~lack of guidelines on implementing both the well-studied network properties in social networks and features, 2)~insufficient research on simulating dynamic social networks with node features, 3)~lack of rigorous study providing directions on defining node labels in social networks, we propose a framework for social graph simulation. In our model, the simulated networks have the following characteristics based on understanding of Facebook-type social networks, along with well-studied social network properties such as preferential attachment. 

\begin{itemize}

    \item Node features are evaluated by other nodes before connecting. If two nodes are forming a connection, the decision of forming a link is taken by both of the nodes, thus both parties should evaluate each other's features. 
    
    \item The decision of forming a connection is based on the preferences of nodes, which are consist of a set of latent variables. These preferences are not directly linked with users' features. For example, two people could live in any state or county, but the preference towards a particular political party could be same, thus resulting in different features but common preferences. 
    
    \item People have common preferences. For example, a group of people in social network may prefer a common ideology or political view. 
    
    \item The node and graph level characteristics should both be taken into account while modelling a network. Node level characteristics consist of features (i.e. node attributes such as age, gender, etc.), individual preferences (latent variables such as preference towards a particular type of people, discussed in more detail in the Section~\ref{our approach}), node degree (i.e. preferential attachment). Whereas, graph level characteristics is e.g. smaller path length preference, i.e. connecting with friends who are nearer in terms of the graph topology.
\end{itemize}

\section{Proposed approach}
\label{our approach}
The proposed simulation is based on preferences (i.e. a set of latent variables) of nodes, which can be interpreted as social rules. Node preference represents the preference of a person in a social network, and at the same time the network-based projection of personality and behaviour. This in turn translates into the network topology. For example, one of network-based behaviours might be to only connect with people with many mutual friends, shaping the topology of the ego network. 

We identify the types of preference a node can have based on their topological and non-topological characteristics. The preferences/behaviours emerge from the following phenomena:

\begin{itemize}

     \item Feature-based (non-topological): From node/user point of view, a combination of variable preferences towards the features of other nodes/users acts as a deciding factor for who they wish to connect with. For example, someone may prefer to connect with people who live near in terms geographic location (e.g. city, town), thus similar location feature is preferred. Whereas, for some other features such as gender, being opposite or same, could be preferred. Thus, for this particular node, preference towards geographic location in combination with gender is considered while connecting.
    
    \item Topology-based: Besides node features, the local topological characteristics may also play important role, e.g. someone may prefer to connect with other people with whom he/she has mutual friends, whereas others may be more open. Secondly, some people may still prefer to connect with someone who has many friends, i.e. popular nodes. Both of these preferences are solely based on the graph topology and could be mostly identified via the topological properties of the social graph.
    
    \item Hybrid feature and topology based (combination of topological and non-topological features): People in social networks may prefer someone who is nearer to them in terms of geographic location and has similar age, education level and has many mutual friends. In this scenario, we have a combination of both the feature-based and topology-based preferences. Someone may also only connect with a politician who has many friends, only if, he or she has similar political views.
    
\end{itemize}

All three types of human preferences are reflected in both non-topological features of a node and the topology of a graph. Although the first, feature-based preference, solely emerges from the non-topological component of a node, once the connections are made, these preferences are also reflected or encoded within the topology of the graph. Without the consideration of the graph or relations between nodes, the predictive model will not be able to capture these complex patterns, which may negatively influence model performance. As a result, by including node attributes in graphs we can achieve higher predictive power.

We propose that the labels of social networks in supervised or semi-supervised classification will capture patterns resulting from the preferences discussed above. We name these preferences or behaviours for a particular node as their social-DNA (\emph{sDNA}). Although most people in a social graph have different features, many people have a similar \emph{sDNA} (i.e. they share preferences). As a result, the \emph{sDNA} is the most valuable and meaningful candidate for class labels for grouping nodes in a graph. However, these labels may not be explicitly defined for a given node classification problem. For example, a classification task may require identifying a group of nodes who may prefer to buy a certain type of product, for marketing purposes. The label for the class who have bought the products should capture a certain group of people with similar preferences in the social network. In semi-supervised classification, if we have a dataset for only a few people who may have bought the product, the classification model would associate a certain type of \emph{sDNA} in the social network as the most likely group to buy that product. In addition, if we do not have any historical information which tells us who have bought the product, it may be possible that a group of people with similar \emph{sDNA} may prefer the product more often than others. However, a person or node itself in a social network, may not entirely know his or her preferences or \emph{sDNA}. As a result, finding these preferences (i.e. labels) in terms of \emph{sDNA} is a nontrivial task. One solution to this problem could be to define a few randomly selected nodes with different labels. These labels could also be selected based on a strategy of focusing on the features of nodes. For example, selecting a few nodes with very different features from each other. After labelling, the semi-supervised classification algorithm, such as GCN will infer other nodes with similar preferences/\emph{sDNA}. Even if no prior knowledge is available, randomly selected nodes with different class labels could be used with only one label per class. GCN is shown to be powerful enough to accurately classify nodes with only one label per class~\cite{kipf2016semi}.    

\section{Graph Formation}
\label{graphFormation}
Let's assume that we have $|N|=n$ number of nodes with $\mathbf{f}$ features each. Each element or feature in the $\mathbf{f}$ could be unbounded or bounded. Each and every node subscribes to an \emph{sDNA}. There is a total of $y$ different \emph{sDNA}s, such that $y<=n$. Nodes which subscribe to the same \emph{sDNA} have the same preference, thus the same label. Each \emph{sDNA} consists of two vectors of length $\mathbf{f}$ (i.e. the same as the number of node features). These two vectors are, 1)~$\mathbf{\tilde{w}}$, which defines the strength or weight of a particular feature's preference and ranges between $0$ and $1$, and 2)~$\mathbf{\tilde{l}}$, which defines whether similar or dissimilar features are preferred with a binary attribute $1$ or $-1$. Although $\mathbf{\tilde{l}}$ could be incorporated into $\mathbf{\tilde{w}}$ as its sign, to make the preference standout separately for user readability and its contribution in the \emph{sDNA} mutation process discussed in Section~\ref{dynamic graph} for dynamic graphs, the vector $\mathbf{\tilde{l}}$ is used. This also allows to have a separate label for the preferences within the \emph{sDNA}, which could be learned using machine learning algorithms from the graph enabling a more in-depth predictive analysis.  
The feature-based scores between two nodes $i$ and $j$ are calculated as (where $\odot$ is the Hadamard product):

\begin{equation}
     \Phi _{i \rightarrow j} = |\mathbf{f}_i-\mathbf{f}_j|^\top (\mathbf{\tilde{w}_i} \odot \mathbf{\tilde{l}_i}) 
    \label{scoreFeatureI}
\end{equation}

Equation~\ref{scoreFeatureI} gives feature-based score which entails if node $j$ is a potential friend for node $i$. In this case, node $i$ evaluates if it wants to connect with node $j$ or not, as we consider only $i$'s \emph{sDNA}. In many social networks, mainly for the undirected ones, the final connection or friendship decision is made by both of the nodes. We can introduce two-way evaluation simply by adding node $j$'s \emph{sDNA} based score in Equation~\ref{scoreFeatureI}. 

\begin{equation}
     \Phi _{i \leftrightarrow j} = \Phi _{i \rightarrow j} + \Phi _{j \rightarrow i} 
    \label{scoreFeatureIJ}
\end{equation}

If both the $i$ and $j$ subscribe to the same \emph{sDNA} then $\Phi _{i \leftrightarrow j} = 2 \Phi _{i \rightarrow j}$. However, Equation~\ref{scoreFeatureIJ} does not prefer similar \emph{sDNA} over different \emph{sDNA} or vice versa. In a social network, the preference or \emph{sDNA} is a set of latent variables. It may well be that two people have a similar preference and this results in a lower score. For example, if two social network users prefer to connect with the opposite gender more often, then if they have the same gender then they are less likely to connect.  

\begin{figure}[h]
\centering
\includegraphics[width=0.5\textwidth]{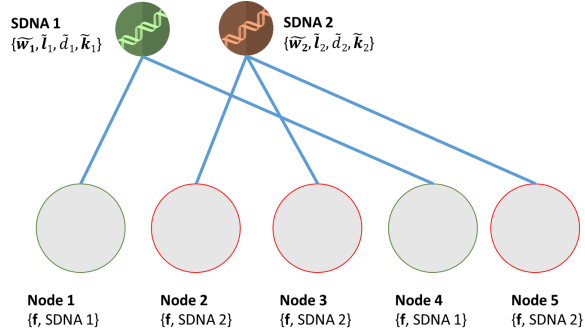}
\caption{Two types of \emph{sDNA} subscribed by 5 nodes (The lines do not represent edges in the graph and \emph{sDNA}s are not nodes. The arrows define subscription or common preferences of different nodes as \emph{sDNA}s)}
\label{sdna}
\end{figure}

Equation~\ref{scoreFeatureIJ} does not consider topological (i.e. graph based geometric features) of the nodes while calculating the score. In social networks, popularity, i.e. the degree of the nodes is a common topological feature with a significant effect on the growth of the network. Typically, people tend to prefer other people who have large number of connections. This is why famous people tend to get more connections. This phenomenon is well studied and known as preferential attachment~\cite{barabasi1999emergence}. To add the preferential attachment effect, we can simply add the degree of the connected nodes to the score. If node $j$ has $m_j$ degree or connections, then from $i$'s perspective, the popularity-based score can be calculated as follows: 

\begin{equation}
    \delta_{i \rightarrow j} = m_j
    \label{scoreDegreeItemp}
\end{equation}

However, the preference for nodes with higher degrees varies from person to person. We can incorporate this variability by including \emph{sDNA}'s preferential attachment parameter while calculating the score, resulting in:

\begin{equation}
  \Delta_{i \rightarrow j} = m_j \tilde{d}_i
  \label{scoreDegreeI}
\end{equation}

Equation~\ref{scoreDegreeI} only considers score of $i$ and $j$ from $i$'s perspective (i.e. $i$'s \emph{sDNA}). For an undirected graph, we can use the following Equation~\ref{scoreDegreeIJ} to calculate preferential-attachment-based score from both $i$ and $j$'s perspective by adding both of their scores from each other's perspective: 

\begin{equation}
  \Delta_{i \leftrightarrow j} =  \Delta_{i \rightarrow j} +  \Delta_{j \rightarrow i} 
  \label{scoreDegreeIJ}
\end{equation}

A social network user tends to prefer people who are nearer to them in terms of the graph topological distance~\cite{wahid2017newton}. Creating a connection with somebody who is friend-of-our-friend is usually more likely than starting a relationship with someone who is further away from us in the structure. However, this preference is also subjective and varies among social network users. As a result, we add this variability of path length preferences subjective to a user by using \emph{sDNA}'s $\mathbf{\tilde{k}} = {\tilde{k}_2,\tilde{k}_3,....\tilde{k}_q} $ variables, where $q$ is the longest path-length in the graph that the model is considering, and $\tilde{k}_2 > \tilde{k}_3 > .... > \tilde{k}_q$. The \emph{sDNA}'s $\mathbf{\tilde{k}}$ vector has a length of $(q-1)$.  

\[
[\mathbf{A}^x[i,j]] =
 \begin{cases} 
      0 & \mathbf{A}^x[i,j] = 0 \\
      1 & \mathbf{A}^x[i,j] > 0 
   \end{cases}
\]

\begin{equation}
\begin{split}
     \mathbf{\mathbf{\Pi}}_{i \rightarrow j}= [\mathbf{A}^2[i,j]]\tilde{k}_(i,2), [\mathbf{A}^3[i,j]]\tilde{k}_(i,3),\\.... ,[\mathbf{A}^n[i,j]]\tilde{k}_(i,q)
    \label{scorePathI}
    \end{split}
\end{equation}

In Equation~\ref{scorePathI}, $[\mathbf{A}^x[i,j]]$ is a generalised Kronecker delta function in the Iverson bracket where $\mathbf{A}$ is the adjacency matrix of the graph. The value of $[\mathbf{A}^x[i,j]]$ is one, if a path of length $x$ between $i$ and $j$ exists, and zero otherwise. This function of path length introduces non-linearity in the score. Equation~\ref{scorePathI} gives the score of $j$ when $i$ is evaluating $j$'s potential to be able to connect or become friends with $i$. This is done by using $i$'s \emph{sDNA} parameter to calculate the score of $J$.  This imitates the behaviour, firstly, $i$ may find someone $j$ interesting to send him a friend request on Facebook. Secondly, the final connection will be made if $j$ also finds $i$ interesting. Equation~\ref{scorePathI} accounts for the first case and the following Equation~\ref{scorePathIJ}, similar to score based on features in Equation~\ref{scoreFeatureIJ} accounts for the score from $j$'s point of view for $i$. The final score function based on path topology is: 

\begin{equation}
     \mathbf{\Pi}_{i \leftrightarrow j}=   \mathbf{\Pi}_{i \rightarrow j} + \mathbf{\Pi}_{j \rightarrow i}
    \label{scorePathIJ}
\end{equation}

Equation~\ref{scorePathIJ} is required if were to simulate a directed graph. For an undirected graph, we only consider Equation~\ref{scorePathI}.  

Finally, we add all three scores, i) feature based from Equation~\ref{scoreFeatureIJ}, ii) popularity based from Equation~\ref{scoreDegreeIJ}, and iii) shortest path length based from Equation~\ref{scorePathIJ} (directed) or ~\ref{scorePathI} (undirected) to calculate the final score $s$. We consider an undirected graph where both $i$ and $j$ make mutual decision to connect with each other. In case of an undirected graph, for $i$ connecting with $j$, we can simply consider Equations~\ref{scoreFeatureI},~\ref{scoreDegreeI}, and~\ref{scorePathI}. For simplicity we are not including the subscript $i \leftrightarrow j$ in the final score function $s$.

\begin{equation}
  s(\Phi,\Delta,\mathbf{\Pi}) =  \Phi + \Delta + \mathbf{\Pi}
  \label{final1}
\end{equation}

Equation~\ref{final1} gives the score between any two nodes. The scores are weighted or modified according to the \emph{sDNA} a node belongs to. To further enforce some global graph level control in the effects of feature-based, popularity-based, and shortest-path scores we introduce two hyperparameters. This global control is useful in many situations, for example, one may wish to generate networks where strong preferential attachment phenomena exist. To be able to control this global weighting, we introduce $r$ and $\mathbf{c}$ global weighting factors in Equation~\ref{final1}. $ \mathbf{c}$ is a vector of length $q-1$, where $q-1$ is the number of shortest path length considered starting from length two. 

\begin{equation}
  s(\Phi,\Delta,\mathbf{\Pi}) =  \Phi + r \Delta +  \mathbf{c}^\top  \mathbf{\Pi}
  \label{final2}
\end{equation}

Equation~\ref{final2} contains $sDNA = \{ \mathbf{\tilde{w}}, \mathbf{\tilde{l}}, \tilde{d}, \mathbf{\tilde{k}} \}$ variables from $\Phi$, $\Delta$, and $\mathbf{\Pi}$. These do not come from the nodes directly but from their \emph{sDNA}s, which in turn expresses their behaviour in the network. 
Equation~\ref{final2} is one possible linear combination of $\Phi$, $\Delta$, and $\mathbf{\Pi}$, however, other possible nonlinear combination functions may be used depending on the target domain.  

\section{Simulation Process}
\label{Simulation Process}
The link formation process for a graph with $n$ nodes is given in Algorithm~\ref{AlgoSocialise}. Each node subscribes to exactly one of $y$ different types of \emph{sDNA} (Figure~\ref{sdna}) and contains $f$ features. In Line~\ref{AlgoSocialise_line_2} the algorithm generates all pairs of nodes. In case of an undirected graph, the pairwise permutation (without repetition) is considered. Furthermore, if self-connection is desired then self pairwise combination is also included. Social network users do not necessarily explore all potential friends whom they might connect with. For example, a Facebook user does not explore all existing Facebook users to connect with. As a result, the simulation process selects a pair of nodes to calculate scores with the exploration probability $p$ (Line~\ref{AlgoSocialise_line_6}), much like how connections are made in a random graph. $p=1$ will result in calculation of scores for all possible pairs, while for $p=0$, no score will be calculated between any pair of nodes. The exploration probability $p$ incorporates controlled stochasticity. In order to determine the minimum score a pair of nodes should have to connect, we define a cut-off point $t$. 


To sum up, first, we calculate scores between pairs selected based on the exploration probability $p$ and then we sort these scores in descending order. After that, we connect $t$ fraction of pairs of nodes in the entire graph. Smaller values of $t$ will result in a social network where the users are very particular about with whom they connect. On the other hand, very high values of $t$ will result in a network where users do not care about features or topological properties while connecting. Thus the latter will be close to a random graph model with probability of edge occurring being equal to $p$, i.e. $t=1$ will result in a pure random graph model with $p$ probability of edges formation.  In Line~\ref{AlgoSocialise_line_4}~and~\ref{AlgoSocialise_line_5}, from all sets of pairs we select a pair for score calculation with probability, $p$. In Line~\ref{AlgoSocialise_line_5}, $r(0,1)$ is a random number generator function which returns a random number from 0 to 1 from uniform distribution. 
In Line~\ref{AlgoSocialise_line_6}, we use Equation~\ref{final2} to calculate a score between the selected pairs of nodes. Afterwards, the stopping length based on the suggested fraction of node pairs to be connected is calculated (Line~\ref{AlgoSocialise_line_11}). In Line~\ref{AlgoSocialise_line_12} we sort the selected pairwise nodes' scores in descending order. Afterwards, in Line~\ref{AlgoSocialise_line_14}, we connect the first $t$ fraction pairs of nodes' for which we have calculated scores in $Scors$, in Line~\ref{AlgoSocialise_line_12}, thus, pairs with higher scores will have a higher likelihood of forming connections.
   
 
 

\begin{algorithm}
\caption{Socialise algorithm}\label{AlgoSocialise}
\begin{algorithmic}[1]
\Procedure{Socialise}{$N,p,t$} \label{AlgoSocialise_line_1}
\State $\mathbf{Pairs} \gets PairCombination(N)$ \label{AlgoSocialise_line_2}
\State $i \gets 0$  \label{AlgoSocialise_line_3}
\ForAll{pair in $\mathbf{Pairs}$}  \label{AlgoSocialise_line_4}
         \If{$r[0,1] \leq p$}  \label{AlgoSocialise_line_5}
              \State $\mathbf{Scores[i]} \gets s(pairs)$  \label{AlgoSocialise_line_6}
              \State $i++$  \label{AlgoSocialise_line_7}
         \EndIf  \label{AlgoSocialise_line_8}
\EndFor \label{AlgoSocialise_line_9}
\State $i \gets 0$ \label{AlgoSocialise_line_10}
\State $StoppingLen \gets \floor{t \times Length(\mathbf{Pairs})}  $ \label{AlgoSocialise_line_11}

\State $\mathbf{Scores} \gets sort(Scores, descending =True)$ \label{AlgoSocialise_line_12}

    \ForAll{score in $\mathbf{Scores}$} \label{AlgoSocialise_line_13}
        \State $Connect(\mathbf{pairs})$ \label{AlgoSocialise_line_14}
        \If{$i \geq StoppingLen$} \label{AlgoSocialise_line_15}
             break 
        \EndIf \label{AlgoSocialise_line_16}
        \State $i++$ \label{AlgoSocialise_line_17}
    \EndFor \label{AlgoSocialise_line_18}
\EndProcedure \label{AlgoSocialise_line_19}
\end{algorithmic}
\end{algorithm}

\section{Curse of Dimensionality in Networks}

\begin{figure}[h]
\centering
\includegraphics[width=0.5\textwidth]{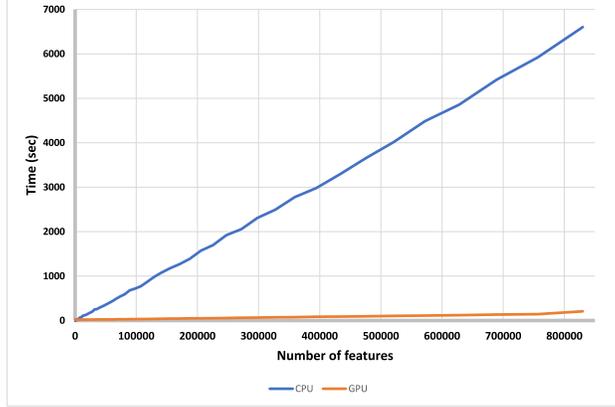}
\caption{CPU vs GPU computation time with varying number of features. (CPU: Intel(R) Xeon(R) W3680 @ 3.33GHz 6 cores and 12 threads, system memory: DIMM DDR3- 20 GB, GPU: NVIDIA GeForce GTX 1080Ti) }
\label{GPUcompute}
\end{figure}

Real-world social networks contain high dimensional features. 
If we consider a Facebook user's posts, likes, photos, comments, etc. as features, then we have thousands of features for each of the node. One problem with nodes with high dimensional features is the linear increase of computational complexity for the simulation process discussed in Section~\ref{Simulation Process}. To overcome this problem, GPU computation can be used to calculate Scores in Equation~\ref{final2}. In our simulation library, we have enabled GPU computation and Figure~\ref{GPUcompute} shows computation time with 300 nodes with increasing number of features. 

\section{Dynamic graph}
\label{dynamic graph}

Real social networks evolve over time and are dynamic in their nature. However, dynamic graph datasets are very rare to find, especially with ground-truth labels and node attributes. These datasets are crucial in the field of dynamic graph research, but also essential for the evaluation of a link prediction, which usually deals only with static graphs or a snapshot of a graph at time $t$. The link prediction problem is to identify new links that will be present in the network at time $t+1$~\cite{bliss2014evolutionary, hristova2016multilayer}. Assuming the network has a set $N$ of nodes and set $E$ of edges at time $t$ expressed as $G(N,E_t)$, and that a link between a pair of vertices $i$ and $j$ is denoted by $L(i,j)$, the goal of link prediction is to predict whether $L(i,j) \in E_{t+1}$, where $L(i,j) \notin E_t$. The prediction is performed by using topological and/or non-topological information about node characteristics and their relationships. Thus, to evaluate or test the performance of a link prediction method, a future snapshot at $t+1$ time is required. Additionally, machine learning based link prediction algorithms require a future snapshot of the network at time $t+1$ as ground truth for training purposes. 
Interestingly, by using multiple runs of Algorithm~\ref{AlgoSocialise} we can already get dynamic graphs, i.e. a future snapshot of the network. Every time we socialise the graph using Algorithm~\ref{AlgoSocialise} containing pairs of nodes which are not yet connected, we will get new connections occurring within the graph. However, this is perhaps not the best simulation of the dynamic nature of real social networks. The reason is that by running Algorithm~\ref{AlgoSocialise} multiple times we are forcing each of the social network users to make consideration and connect with people which they didn't find interesting enough in the previous run(s)\footnote{Here, we are assuming no arrival of new nodes or a constant number of nodes. In case of new nodes, we can easily run Algorithm~\ref{AlgoSocialise} with the new nodes and include them in the graph.}. What we really want is not to force the users in the graph to make new connections but allow the users' interest and preference to change and then run the Socialise Algorithm~\ref{AlgoSocialise}. This will result in a concept drift in the user preferences, which can be achieved via changing values of the variables in the \emph{sDNA}'s of the nodes. This changes in the \emph{sDNA} reflect the phenomenon that, the rules that govern social networks can and do change over time. This change of preference can be achieved via the \emph{sDNA} Mutation given in Algorithm~\ref{AlgoMutation}. The intensity of mutation can be controlled by mutation intensity parameter $z$, which results in changing values of the variables in
 $sDNA = \{\mathbf{\tilde{w}}, \mathbf{\tilde{l}}, \tilde{d}, \mathbf{\tilde{k}} \}$. A lower value of $z$ would change only few of the $\mathbf{\tilde{w}}$. As a result, the user's preference towards a potential friend's feature $\mathbf{f}$ would change. In case the value of the mutation intensity parameter $z$ is defined larger, this would result in changes to the entire preference vector $\mathbf{\tilde{l}}$. In Algorithm~\ref{AlgoMutation}, in Line~\ref{AlgoMutation_line_1}, the procedure takes all $y$ existing $sDNAs$ from the graph, and $mutatePreference$, a boolian parameter to determine if $l$ should also be changed. In Line~\ref{AlgoMutation_line_2} we iterate trough each of the $sDNAs$, one at a time. For the given \emph{sDNA}, we then iterate trough each of the elements in $\mathbf{\tilde{w}}$ and $\mathbf{\tilde{l}}$ in Line~\ref{AlgoMutation_line_3}. We than reassign the value of $\tilde{w}$ with the probability $z$ (Line~\ref{AlgoSocialise_line_5}). In Line~\ref{AlgoMutation_line_7} we check if the $mutatePreference$ is set True. If so, then we also reassign the value of $\tilde{l}$, $1$ or $-1$, with a probability of $z$ (Line~\ref{AlgoMutation_line_9}). The selection between $1$ or $-1$ selected randomly from uniform random distribution. In Line~\ref{AlgoSocialise_line_14} we reassign the $\tilde{d}$ parameter of \emph{sDNA}, which is for preferential attachment strength. Afterwards, in Line~\ref{AlgoSocialise_line_16}, $q$ number of random numbers are generated, for each path length preference in Equation~\ref{scorePathI}. The intervals are selected such that it satisfies, $\tilde{k}_2 > \tilde{k}_3 > .... > \tilde{k}_q$. Afterwards, in Line~\ref{AlgoMutation_line_20}, we again iterate through each elements of $\mathbf{\tilde{k}}$ and reassign from the already generated random numbers in Line~\ref{AlgoMutation_line_16}.
 
An interesting observation is, generally people's behaviours or preference changes are correlated with time. This change in behaviour, for social network users, contributes to change in the topology of the social network. In our simulation strategy, a snapshot to snapshot time difference then should also be correlated with the change of the users' behaviour or preferences, i.e. \emph{sDNA}s. The parameter $z$ in Algorithm~\ref{AlgoMutation} defines this intensity of mutation in \emph{sDNA} or intensity of social network users change in behaviour. As a result, the value of $z$ is proportional to the time between two snapshots of the network. For example, if one wishes to run the Socialise algorithm (Algorithm~\ref{AlgoSocialise}), it will produce a social network with the first snapshot, $snapshot-1$. Then running the Mutation algorithm (Algorithm~\ref{AlgoMutation}) will result in change in preferences with a particular value of the parameter $z$, and then rerunning the Socialise algorithm will result in another snapshot of the network in a forward time dimension, $snapshot-2$. A high value of $z$ will result in higher time difference between these two snapshots, $snapshot-1$ and $snapshot-2$. 

One may wish to generate event based dynamic networks, i.e. time-stamped link formation. This can also be achieved by setting the `fraction of nodes to be connected' parameter $t$ from the Socialise Algorithm~\ref{AlgoSocialise} such that only one link is formed. A repeated run of Algorithm~\ref{AlgoSocialise} will result in edge stream with timestamp for each of the edges. As we have discussed earlier in the section, the time between each of the edge appearing could also be manipulated by changing the value of parameter $z$.


\begin{algorithm}
\caption{Mutation algorithm}\label{AlgoMutation}
\begin{algorithmic}[1]
\Procedure{Mutate}{$z,\mathbf{sDNA},mutatePreference$} \label{AlgoMutation_line_1}
\ForAll{$sDNA$ in $\mathbf{sDNA}$}\label{AlgoMutation_line_2}
\ForAll{$\tilde{w}$, $\tilde{l}$ in $\mathbf{\tilde{w}}$,$\mathbf{\tilde{l}}$}\label{AlgoMutation_line_3}
         \If{$r(1) \leq z$}\label{AlgoMutation_line_4}
             \State $\tilde{w} \gets r(1)$\label{AlgoMutation_line_5}
         \EndIf\label{AlgoMutation_line_6}
          \If{$mutatePreference$}\label{AlgoMutation_line_7}
             \If{$r(1) \leq z$} \label{AlgoMutation_line_8}
             \State $\tilde{l} \gets rand(-1|1)$ \label{AlgoMutation_line_9}
              \EndIf \label{AlgoMutation_line_10}
         \EndIf\label{AlgoMutation_line_11}
\EndFor \label{AlgoMutation_line_12}
    \If{$r(1) \leq z$}\label{AlgoMutation_line_13}
             \State $\tilde{d} \gets r(1)$\label{AlgoMutation_line_14}
    \EndIf\label{AlgoMutation_line_15}
    \State $K\_r \gets  r[1,\frac{q-1}{q}),  r[\frac{q-1}{q},\frac{q-2}{q}), ..., r[\frac{1}{q},\frac{q-q}{q})$\label{AlgoMutation_line_16}
    \State $i \gets 0$\label{AlgoMutation_line_17}
    \ForAll{$\tilde{k}$ in $\mathbf{\tilde{k}}$}\label{AlgoMutation_line_18}
    \If{$r(1) \leq z$}\label{AlgoMutation_line_19}
             \State $\tilde{k} \gets K\_rand[i]$\label{AlgoMutation_line_20}
             \State i++\label{AlgoMutation_line_21}
    \EndIf\label{AlgoMutation_line_22}
    \EndFor\label{AlgoMutation_line_23}
\EndFor \label{AlgoMutation_line_24}
\EndProcedure\label{AlgoMutation_line_25}
\end{algorithmic}
\end{algorithm}




\section{How to validate simulation}
\label{Validation}
In order to assess if the desired integration of features, labels, and topology is achieved, we measure and compare different trained model's predictability of the labels of the nodes. This comparison is done by designing different setups of the models such that, the models are able to perform predictions with the entire set of information (features, labels, and topology) as well as with partial information. 


Here we discuss the validation setup. The predictability of the label of a node, i.e. \emph{sDNA}, can be performed via the following configurations of an ideal machine learning model:

\begin{enumerate}
  \item Predictability of nodes' \emph{sDNA}s with features combined with the graph topology 
  \item Predictability of nodes' \emph{sDNA}s using features only
  \item Predictability of node's \emph{sDNA} using topology only
\end{enumerate}

We can expect for an ideal machine learning model to fully capture and learn patterns both from the topological and feature based information from the network without over-fitting or being susceptive to the noise or stochasticity in the network. Needless to say, such an ideal model is not currently available in the real-world. However, we should at least use a machine learning model which can directly utilise both the topological and non-topological information, i.e. features.  

In our case we use the GCN~\cite{kipf2016semi} to analyse \emph{sDNA} predictability of the simulated networks, which can be regarded as one of the best models to directly combine both the topological and non-topological information of the graph~\cite{li2018deeper}. 

\subsection{Graph Convolutional Networks (GCNs)}
GCN is a multi-layer graph based neural network. In each layer, the features are multiplied with the topology of a graph in the spectral domain (i.e. symmetric normalised Laplacian matrix~\cite{kipf2016semi}). Weights of connections (edges/links by which the features of a node are passed, considered or summed) are learned using backpropagation. However, as most of the real-world social networks are not regular graphs, one single weight is learned for all links of a particular node. 

The layer-wise propagation rule for the $l$'th layer is: 

\begin{equation}
H^{(l+1)} = \sigma ( GH^{(l)}W^{(l)})
\label{eq:gcn}
\end{equation}

In Equation~\ref{eq:gcn}, $W^{(l)}$ are the trainable weight matrices for each layer. $H^{(0)}=X$ (the feature matrix) and $G$ is a graph representative matrix that we discuss in more detail in Section~\ref{Graph representatives for GCN}. $G$ is fed in every layer of the model until the output layer. Finally, $\sigma$ denotes a nonlinear activation function.

For this model, the receptive field grows with the depth of the network~\cite{kipf2016semi}. In the first layer, only friends' features are considered, and in the second layer friend of friends' features are also considered, i.e. summed before passing through a non-linearity. This is because the summarised friends' information is already gathered in the first layer. 

The direct translation from a graph to the structure of the neural network\footnote{In this paper when we talk about a graph (i.e. a social network) we mention it as a `graph' or a `network' but when we talk about a neural network it is written in its full form.} is achieved via the graph representative matrix $G$.  Symmetric normalised Laplacian matrix of the adjacency matrix $A$ has been used in the original formulation of GCN, i.e. $G=\tilde{L^{sym}}$ ~\cite{kipf2016semi}. 

\begin{equation}
\tilde{L}^{sym} =  \tilde{D}^{-\frac{1}{2}}\tilde{A}\tilde{D}^{-\frac{1}{2}}
\label{eq:laplacian}
\end{equation}

\begin{equation}
\tilde{A} = A + I_N
\label{eq:self-connections}
\end{equation}

In Equation~\ref{eq:self-connections}, $I_N$ adds the self-connections for each of the nodes in $A$, $\tilde{D}$ is the degree matrix of the adjacency matrix, and $\tilde{A}$ is the adjacency matrix with added self-connections. The addition of self-connections facilitates incorporation of self-features of the nodes for better predictability. For example, a social network user's friends may give away his or her preference or class label (i.e. predictability based on the labels of the connected nodes), but additionally, his or her own features (i.e. self-connections in the graph) are also important to consider to predict his or her preference. 


In Equation~\ref{eq:gcn}, the main transformation to the neural network from a graph is performed through $G=\tilde{L^{sym}}$. If the adjacency matrix $A$ in $\tilde{L^{sym}}$ (Equation~\ref{eq:laplacian}) is replaced with a different representative function of the graph, the structure of the neural network itself will change. However, this does not change the input feature matrix $X$. As a result, this is not exactly data preprocessing technique but rather a change in the architecture of the neural network. We discuss this usage of different graph representatives later in Section~\ref{Graph representatives for GCN}.

Using the GCN we calculate the three mentioned setups for node label predictions in Section~\ref{Validation} by changing the propagation rule in Equation~\ref{eq:gcn} as follows: 

\begin{enumerate}
\label{prop_rules_test_simulation}
  \item Prediction of nodes' \emph{sDNA} with both the features and graph topology using propagation rule of Equation~\ref{eq:gcn} for the first layer, where $G=\tilde{L}^{sym}$, and $H^{0}=X$, where $X$ is the feature matrix: 
  
    \begin{equation}
    H^{(1)}_{(\Phi,\Delta,\mathbf{\Pi})} = \sigma ( \tilde{L}^{sym}XW^{(0)})
    \label{eq:gcnVanillaRule1}
    \end{equation}
   This is the straightforward GCN model proposed by~\citet{kipf2016semi}. Here, the graph representative $G=\tilde{L}^{sym}$ is fed in every layer of the model, but the feature matrix $X$ is fed only in the first layer.
    
  \item Prediction of nodes' \emph{sDNA} with features excluding graph topology with the following propagation rule: 
      
      \begin{equation}
    H^{(1)}_{(\Phi)} = \sigma ( I_A XW^{(0)})
    \label{eq:gcnFeatureOnly}
    \end{equation}
   
   In this first layer propagation rule in Equation~\ref{eq:gcnFeatureOnly}, $I_A$ is the identity matrix of the adjacency matrix $A$. $I_A$ is fed into the model until the output layer. Thus, only features of each node are considered and the graph topology does not play any role for label or \emph{sDNA} predictions. 
   
  \item Prediction of node's \emph{sDNA} excluding the features but solely with the graph topology:
  
   \begin{equation}
    H^{(1)}_{(\Delta,\mathbf{\Pi})} = \sigma (\tilde{L}^{sym} I_X W^{(0)})
    \label{eq:gcnTopologyOnly}
    \end{equation}
   
   In this propagation rule in Equation~\ref{eq:gcnTopologyOnly}, $I_X$ is the identity matrix of the feature matrix $X$. As a result, only the graph topology is considered, and features do not play any role in the model. Here, the graph representative $G=\tilde{L}^{sym}$ is fed in each layer of the model until the output layer of the model, however $H^{0}=I_X$ is fed only in the first layer of the model.  
   
\end{enumerate}

We assume that during the simulations, the first setup will produce more accurate results than the remaining two. This hypothesis is represented through the following inequality:

\begin{equation}
\begin{aligned}
   \Big( Acc(H_{(\Phi,\Delta,\mathbf{\Pi})})> Acc(H_{(\Phi)}) \Big) \land \Big( Acc(H_{(\Phi,\Delta,\mathbf{\Pi})}) > \\ \big ( Acc(H_{(\Delta,\mathbf{\Pi})}) \lor Acc(H_{( \overline {\Delta,\mathbf{\Pi})}}) \big)  \Big)
    \label{inequality}
    \end{aligned}
\end{equation}
where $Acc$ is the test accuracy of the trained neural network model for four different setups, based on four different propagation rules in Equations~\ref{eq:gcnVanillaRule1},~\ref{eq:gcnFeatureOnly}, \ref{eq:gcnTopologyOnly}, and \ref{eq:gcnTopologyOnly2}. 
However, the assumption of GCN is that for a $l^{th}$ layer of the model, only the $l^{th}$ order neighbourhood nodes are influential~\cite{kipf2016semi,li2018deeper}. To work around this problem, we develop a strategy of replacing the adjacency matrix $A$ in the Laplacian transformation in Equation~\ref{eq:laplacian} graph representative function $G$, with three different existing node-similarity measures. In social networks, not all connected nodes have the same influence and in fact, some non-directly connected nodes in the graph may have greater influence over a node in question than the directly connected ones. As a result, usage of the adjacency matrix $A$ as a graph representative $G$ may not always entail the best performance of the neural network. 


\subsection{Node-similarities as Graph Representatives for GCN}
\label{Graph representatives for GCN}
In social networks, the adjacency matrix represents direct links between nodes. In GCN the features propagate through those links. Thus, a node's label is predicted by utilising patterns on the surrounding connected nodes' features and labels. However, in social networks, not all the connections of a given node have same or even similar effect on this node. It can be assumed e.g. that the influence that one node has on its neighbour will increase with the number of their mutual friends. In a similar way, it may happen that  a friend of a friend of node $i$ can influence node $i$ more than a directly connected node (a not influential node, e.g. does not have any common friend with the node $i$). As a result, this effectively changes the representation of the network so one can incorporate these relationship characteristics as a form of social node-similarity-based matrix for the GCN. One way to extract and represent these types of social relationship (not necessarily direct ones) strengths and other information between nodes is to use a matrix which describes the similarity between nodes instead of an adjacency matrix. For example, the Katz similarity measurement considers the number of all direct paths from node $i$ to $j$~\cite{katz1953new}. Thus, more mutual friends would result in a higher number of paths, resulting in a higher value of the Katz score. In this study, we replace the adjacency matrix $A$ with the three different types node-similarity matrices, $\hat{A}$ as they encompass richer information about underlying structure than traditional adjacency matrix. Following are the three node-similarity measures we have considered:


\begin{itemize}

\item \textbf{Katz,} which considers the number of all the paths from node $i$ to $j$~\cite{katz1953new}. The shorter paths have bigger weight (i.e. are more important), which is damped exponentially with the increase of the path length and the $\beta$ parameter ($A$ is the adjacency matrix):

\begin{equation}
Similarity(i,j) = \beta A+\beta \textsuperscript{2} A\textsuperscript{2} +\beta \textsuperscript{3} A\textsuperscript{3}+ \cdot \cdot \cdot
\label{katzSimilarity}
\end{equation}

The above similarity in Equation~\ref{katzSimilarity} will result in the following graph representative $G_{katz}=\tilde{L}^{sym}_{katz}$

\begin{equation}
\tilde{L}^{sym}_{katz} =  \tilde{D}^{-\frac{1}{2}}\tilde{A}_{katz}\tilde{D}^{-\frac{1}{2}}
\label{katzLaplacian}
\end{equation}

\begin{equation}
\tilde{A}_{katz} =  \hat{A}_{katz} + I_N
\label{katzLaplacian self-connections}
\end{equation}

\item \textbf{Rooted PageRank (RPR)} is used by search engines to rank websites. In graph analysis it ranks nodes by the probability of each node being reached via random walk on the graph~\cite{brin2012reprint}. The $Similarity(i,j)$ is calculated using the stationary probability distribution of the degree matrix $D$ in a random walk. The random walk returns to $i$ with probability $\alpha$ at each step, moving to a random neighbour with probability $1-\alpha$. This results in the following graph representative $G_{RPR}=\tilde{L}^{sym}_{RPR}$:

\begin{equation}
\tilde{L}^{sym}_{RPR} =  \tilde{D}^{-\frac{1}{2}}\tilde{A}_{RPR}\tilde{D}^{-\frac{1}{2}}
\label{RPRLaplacian}
\end{equation}

\begin{equation}
\tilde{A}_{RPR} = \hat{A}_{RPR} + I_N
\label{RPRLaplacian self-connections}
\end{equation}

\item \textbf{Graph Gravity (GG),}
Inspired by the Newton's law of universal gravitation, this node-similarity measure uses degree centrality as the mass of the nodes, while the lengths of shortest paths between them act as distances~\cite{wahid2017newton,wahid2019predict}. The above analogy leads to the following formula for calculating the score between two nodes:

\begin{equation}
Similarity(i,j)= \frac{C_D(i) \times C_D(j)}{SP(i,j)^2},
\label{newton}
\end{equation}
where $C_D$ denotes the degree centrality, $SP$ is the shortest path. Node-similarity in Equation~\ref{newton} will result in the following graph representative $G_{GG}=\tilde{L}^{sym}_{GG}$:

\begin{equation}
\tilde{L}^{sym}_{GG} =  \tilde{D}^{-\frac{1}{2}}\tilde{A}_{GG}\tilde{D}^{-\frac{1}{2}}
\label{GGLaplacian}
\end{equation}

\begin{equation}
\tilde{A}_{GG} = \hat{A}_{GG} + I_N
\label{GGLaplacian self-connections}
\end{equation}

\end{itemize}

For all the above three node-similarity measures, each $\hat{A}$, $L2$ represents the nodes-similarity matrix (only for all possible links) which has been preprocessed and reconfigured further which is discussed in Section~\ref{ak_preprocessing}. 




\subsection{Weighted Feature Matrix}
\label{strnght_mat}
$GCN$ is a powerful model for node classification, and it has been shown to perform well even only with the graph topology, i.e. without the feature matrix~\citet{kipf2016semi}. The reason for such a good predictability without the features could be due to two reasons. Firstly, when our focus is on node classification for graph-structured datasets, the preferred features of the nodes should be reflected in the topology of the graph as we have discussed in our simulation process in Section~\ref{graphFormation}. The fact that these features are encoded in the topology may result in a good predictability even when the features are not directly considered in the model. Additionally, this better predictability based on feature only or topology only may vary from node to node. For some nodes, the topology only may have better predictability when compared with the node's feature. This could be due to the fact that topological position of a node overshadows the importance of the features. 


Secondly, the good performances solely based on topology could be because, similar to real-social network users, we have defined our \emph{sDNA} for nodes such that it results in some of the features of other nodes being preferred and some others not (Section~\ref{graphFormation}).
In other words, not all the features play similar roles when it comes to the predictability of the \emph{sDNA}. As a result, in the entire graph, some of the features may be disliked or not preferred by the majority of the nodes when forming graph connections. This is why an additional learnable common weight for a particular feature for all the nodes may result in better predictability. In our analysis, we have found that adding this additional weight, which defines the weight for each of the features for all the nodes, seems to perform best, and this is what we present in Section~\ref{results}.
To introduce this relative importance of features we use one additional weight vector in the $GCN$ model. We use a common weight for a particular feature for all the nodes. If we have a network with $1,000$ nodes and $50$ different features each, for each feature of all the $1,000$ nodes a common (i.e. across all the nodes) weight is used to learn the strength of each feature. This additional feature weight matrix is the size of the number of features and is used only in the first layer of the model. Hence all the input features, $X$ are weighted before passing to the hidden layers.    

This additional weight vector results in the following first layer propagation rule based on the Equation~\ref{eq:gcn}:


\begin{equation}
H^{(2)} = \sigma (G ((\mathbf{1} S) \odot X)W^{(1)})
\label{eq:gcn_feat_weight_mat}
\end{equation}
where $S^{1 \times|f|}$ is the matrix containing the unbounded learnable parameters to define strength of the feature matrix $X^{|N|\times|f|}$. $1^{1\times|f|}$ is an all one matrix. $\odot$ defines the Hadamard product between the feature matrix $X$ and the dot product of $1$ and $S$.

\section{Experimental Setup}
In the experiments, we simulate $30$ social networks, with $1,000$ nodes each. Each of the networks has four different types of \emph{sDNA}s with $250$ nodes subscribing to a single type. We take three different snapshots of the same network, resulting from an initial $10$ networks to a total of $30$ networks (i.e. three snapshots of the same network). Each of the nodes has $50$ features, each set of features of a node is generated from a uniform distribution. All the variables for the \emph{sDNA} (described in Figure~\ref{sdna} and Section~\ref{graphFormation}) are also generated from uniform distributions.
For all the models, we have four graph convolutional layers. All layers except the output layer use rectified linear units (ReLU) as nonlinear activation functions. The output graph convolutional layer contains softmax activation and categorical cross entropy loss is calculated for the four types of \emph{sDNA}s or node labels. Each of the layers, except for the output layer contains $32$ units of neurons, and the output layer has $1,000$ units, the same as the number of nodes that need to be classified. Finally, we used Adam optimiser, a first-order gradient-based algorithm for our differentiable neural network model to learn the weights (i.e. to optimise the loss function). Each and every model is evaluated with the same setup. On every network, 10-fold cross-validation is performed and the average accuracy is reported. Additionally, the standard deviation of the accuracy is reported for the best accuracy and the original GCN in Table~\ref{tab:bestMethods}.
All the hyperparameters are kept fixed for all the models. We have used the learning rate of $0.01$, L2 kernel regularisation  (i.e. weight decay) for all the hidden layers with the decay rate of $0.0005$, and a dropout layer after each hidden layer with  $p=0.5$, i.e 50\% of the randomly selected neurons are trained in each training iteration. 
 
\renewcommand{\arraystretch}{1.6}
\begin{table}[!h]
\begin{tabular}{|l|l|l|l|}
\hline
Model     & Description                   & Graph representative $G$          & Eq.                       \\ \hline
FTVanilla & Original GCN, feature + toplogy & $G=\tilde{L^{sym}}$               & ~\ref{eq:gcn}             \\\hline
T         & Original GCN, topology only     & $G=\tilde{L^{sym}}$               & ~\ref{eq:gcnTopologyOnly} \\\hline
TLR         & Original GCN, topology only     & $G=\tilde{L^{sym}}$               & ~\ref{eq:gcnTopologyOnly2} \\\hline

F         & Original GCN, feature only      & N/A                               & ~\ref{eq:gcnTopologyOnly} \\\hline
FTKatz    & Feature + Katz based toplogy        & $G_{katz}=\tilde{L}^{sym}_{katz}$ & ~\ref{katzLaplacian}      \\\hline
FTRPR     & Feature + RPR based toplogy         & $G_{RPR}=\tilde{L}^{sym}_{RPR}$   & ~\ref{RPRLaplacian}       \\\hline
FTGG    & Feature + GG based toplogy        & $G_{GG}=\tilde{L}^{sym}_{GG}$   & ~\ref{newton}  \\ \hline         
\end{tabular}
\smallskip
\caption{Models used along with the original GCN. All of the models with features are trained twice, once with the weighted feature matrix in Equation~\ref{eq:gcn_feat_weight_mat} and once without}
\label{tab:setup}
\end{table}

In Table~\ref{tab:setup}, for the topology only model, \emph{T} (Equation~\ref{eq:gcnTopologyOnly}), the weight matrix contains more trainable parameters compared with the the model in \emph{FTVanilla} (Equation~\ref{eq:gcnVanillaRule1}). This is because we have $1000$ nodes per network with $50$ features each. As a result for the model with both the topology and feature matrix model, \emph{FTVanilla} the dimension of the first layer weight matrix $w^{(0)}$ needs to be $50 \times 32$, where $32$ is the hyperparameter for the number of units we consider in all the models, i.e. $\tilde{L}^{{sym, 1000 \times 1000}} X^{1000 \times 50} W^{ (0), 50 \times 32}$, and the resulting matrix has a dimension of $1000 \times 32$, while the output from the first layer has a dimension of $1000 \times 32$. Whereas for the topology only \emph{T}, in Equation~\ref{eq:gcnTopologyOnly}, where the feature matrix is only an identity matrix, $I_X$, the weight matrix $w^{(0)}$ is directly multiplied with the graph representative, i.e. the graph topology, $\tilde{L}^{{sym, 1000 \times 1000}}$. As a result the dimension of the weight matrix is a lot higher ($1000 \times 32$), i.e. $\tilde{L}^{{sym, 1000 \times1000}}W^{{(0), 1000 \times 32}}$ , and the resulting output from the first layer has a dimension of $1000 \times 32$. As we can see there are more trainable parameters in the \emph{T} model compared with \emph{FTVanilla}, i.e. $50 \times 32$ vs $1000 \times 32$. If we were to compare both of the models' performance, \emph{T} vs \emph{FTVanilla}, to test if the Inequality~\ref{inequality} holds as a validation of the feature and topology integration process, the total number of trainable parameters for the both the models should be as close as possible. To make both the models comparable, we introduce another setup for the topology only model to keep the number of parameters at the similar level to the model using both the features and topology.

\begin{equation}
  H^{(1)}_{( \overline {\Delta,\mathbf{\Pi})}} = \sigma (\tilde{L}^{sym} I_X W^{(a0)} W^{(b0)})
    \label{eq:gcnTopologyOnly2}
\end{equation}
 
In Equation~\ref{eq:gcnTopologyOnly2}, weight matrix $W^{{(0), 1000 \times 32}}$ is split into two matrices $ W^{(a0)_{1000 \times 1}}$ $W^{(b0)_{1 \times 32}}$ to keep the number of trainable parameters roughly in line with the \emph{FTVanilla} model.




\subsection{Augmented node-similarity matrix}
\label{ak_preprocessing}

For $\hat{A}_{GG}$ (Equation~\ref{GGLaplacian self-connections}), the similarity scores for all the non-existing links are calculated and then all the scores are normalised between zero and one. Afterwards, the adjacency matrix $A$ is summed with the calculated scores for all the non-existing links. As a result, all the existing links for $\hat{A}_{GG}$ has a value of one and for the non-existing links, the value ranges from zero to one. For $\hat{A}_{katz}$ and  $\hat{A}_{RPR}$, the path-based similarity scores are calculated for all possible links. For all the networks, to calculate Katz score, with the highest exponent of five for the adjacency matrix $A$ (i.e. $A^5$ in Equation~\ref{katzSimilarity}) and the $\beta=0.005$ is used.  As for the RPR, the $\alpha$ parameter is set to $0.85$. For each of the calculated similarity matrices ($\hat{A}_{GG}$, $\hat{A}_{katz}$ and  $\hat{A}_{RPR}$ ), the row is normalised for each of the non-zero elements using the $L2$ norm. Moreover, on the similarity-based adjacency matrices (i.e.  $\hat{A}_{katz}$,  $\hat{A}_{RPR}$, and  $\hat{A}_{GG}$), several thresholds are used. The thresholds are applied on the $L2$ row normalised matrices. The thresholds are set in a way that, if the value in the similarity-based matrix is less than or equal to the first threshold then it is set to zero. Whereas for the second threshold point, if the value is greater than the threshold, it is set to one. If the thresholds are set as zero and one respectively, then none of the values is changed in the matrix. Also, for some set of thresholds, if they are not the same, the elements in the matrix which are in between the two thresholds, are unaltered in the matrix. The sets of thresholds are selected based on empirical analysis, i.e. cross-validation accuracy of the model. However,  we also select a threshold based on the mean value of the elements of the matrix. The mean value threshold hold is applied such that, if a non-zero element in the matrix is less than or equal to the mean value then it is set to zero and one otherwise. 

In GCN, for the $l^{th}$ layer, only the $l^{th}$ path length neighbouring nodes are considered~\cite{kipf2016semi}. Thus, it limits the scope of the receptive field of the node in each layer and also the maximum receptive field is limited by the maximum number of layers used in the model. This limitation has also been pointed out in the paper where GCN was first introduced~\cite{kipf2016semi}. 
However, using node-similarity measures along with the augmentation process we describe here allows the model to consider a three-path distant node $j$ even in the first layer (i.e. a direct connection) for the classification of the node $i$, assuming that they have a high node-similarity score. As a result, this augmented node-similarity measure solves the limitation of layer-wise node-neighbourhood dependencies for the GCN.

\section{Results and Discussion}
\label{results}
\begin{figure}[H]
    \centering
\includegraphics[width=18cm,height=19cm,trim={0cm 0 0 0},clip]{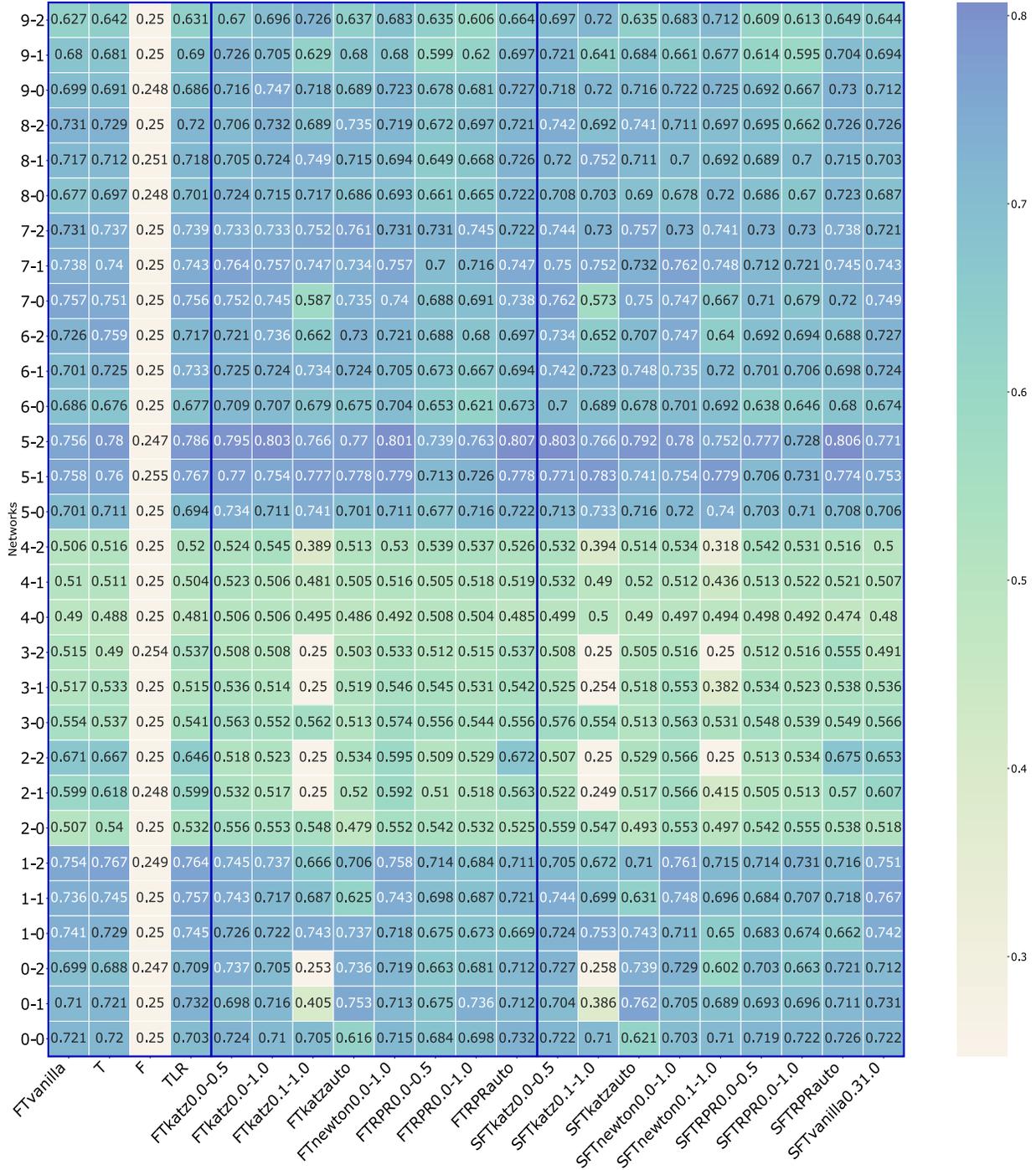}
    \caption{Accuracy from different Models (average from 10-fold cross-validation). Models written as, $F$- feature only, $T$ - topology only, $FT$ - both the feature and topology, $FTvanilla$ - the original GCN. Models with $S$ in the right box represents if an additional feature weight matrix in the first layer (Equation~\ref{eq:gcn_feat_weight_mat}). The left box shows results for models that use graph representative $G=\tilde{L^{sym}}$, Equation~\ref{eq:gcn} (i.e. adjacency matrix). The middle box uses $G_{NS}=\tilde{L}^{sym}_{NS}$, where $NS$ is a node-similarity ($Katz$, $RPR$, and $GG$) measure with different thresholds ( Equation~\ref{katzLaplacian},~\ref{RPRLaplacian} and ,~\ref{newton}). Similarity-based $G$ is preprocessed based on Section~\ref{ak_preprocessing}. The preprocessing threshold $auto$ implies automatic selection of a threshold based on the mean value of the $\hat{A}$ (Section~\ref{ak_preprocessing}). All the networks are represented in terms of snapshots. For example, 0-0, is the first network's first snapshot, 0-1 is the first network's second.}
    \label{fig:results}
\end{figure}

In Figure~\ref{fig:results}, we show accuracy for all the models that we have tested on 30 simulated networks. All the results are 10-folds cross-validated and average accuracy is reported. In Figure~\ref{fig:results} and Table~\ref{tab:bestMethods}, we observe that according to the hypothesis of Equation~\ref{inequality}, the accuracy of the model which uses node features only, i.e. $F$, is very low. In fact, the predictability is not better than random chance (the accuracy is around $0.25$ and we have four equally represented labels or \emph{sDNA} types to predict). Additionally, from Figure~\ref{fig:results} and Table~\ref{tab:bestMethods} we can see that for the majority of the datasets (except only three networks) models which utilise both the topology and features of the graph perform better than the two other setups where topology and features are considered independently. When only topology is used (i.e. T), the model T performs the best in three networks. Two of them are a third snapshot(i.e. the 3rd run of the Algorithm~\ref{AlgoSocialise}) of a network (networks 1-2 and 6-2), and the third one is the second snapshot of the network (2-1). This can be due to the fact that as we run Algorithm~\ref{AlgoSocialise} multiple times, the patterns of preferences get encoded within the network topology  so that the topology only model performs better. This is something we also expect in real world networks i.e. as a person makes more connections, their connection making patter becomes eminent.  

Amongst methods using node-similarity matrices instead of the node adjacency matrix, we see that the choice of threshold seems to have a significant effect on the model-performance. However, we can also see that using the mean value of the L2 normalised node-similarity matrix as a threshold (described in Section~\ref{ak_preprocessing}) performs quite well. In fact, on seven networks with the setup of using mean value as a threshold on the node-similarity matrix (discussed in Section~\ref{ak_preprocessing}) outperforms all the other models (Table~\ref{tab:bestMethods}). The models with the mean value as thresholds are written as `auto' in Table~\ref{tab:bestMethods}.

If we do not consider differences between thresholds and usage of the vector $S$ on the node-similarity measures, $Katz$ significantly outperforms the original GCN (i.e. $FTvanilla$) and two other node-similarity measures ($RPR$ and $GG$) based on the results in Table~\ref{tab:bestMethods}. On five networks, $RPR$ performs the best and $GG$ on one network. However, on the basis of average best-performing models on all the datasets from all the $20$ different models, following models performs best: 1) $FTGG0.0-1.0$,  2) $SFTkatz0.0-0.5$, and 3) $FTkatz0.0-0.5$. Thus, on average, $GG$ performs best across all the datasets. 

The results also show that, the usage of a trainable parameter $S$ based on Equation~\ref{eq:gcn_feat_weight_mat} gives us a better model for many datasets than when not using it. In 15 out of 30 datasets, using $S$ on the first layer of the model outperforms the other models (Table~\ref{tab:bestMethods}). Furthermore, models with $S$ which perform best are mainly not the original GCN but the node-similarity-based models, except for one dataset. However, this may not imply that the use of additional weights in the first layer based on Equation~\ref{eq:gcn_feat_weight_mat} only performs well on node-similarity-based models. This is because the usage of node-similarity may have better predictability in general than the adjacency matrix. 

From Figure~\ref{fig:results} and Table~\ref{tab:setup} we can see that the performance of a node-similarity-based model varies depending on the network the model is trained on. This is because all the networks are simulated with different rules, and no two networks are exactly the same. We can expect to see the same in real-world networks as well. Thus, the choice of a node-similarity method could be based on empirical analysis. However, one may also use the mean value of the normalised node-similarity matrix, especially with GG as we have discussed earlier in this section. 

The results in Figure~\ref{fig:results} show that we can achieve high accuracy (in fact higher than using only the adjacency matrix) on node classification when a node-similarity-based graph topology is used. This is particularly useful for very dense networks. The training time that is required for a very dense network is extremely high for GCN. Many real-world datasets, such as face-to-face interaction networks, tend to be very dense. Thus, the similarity-based matrices can be used (with a suitable threshold to reduce the number of connections as per Section~\ref{ak_preprocessing}) in such scenario to reduce training time.

\begin{table}[H]
\begin{tabular}{|l|l|l|l|}
\hline
Networks & FTvanilla(\cite{kipf2016semi}) Acc (SD) & Max Acc (SD)& Max Acc Model    \\\hline
0-0      & 0.721 (0.011)   & 0.732 (0.007)   & FTRPRauto      \\\hline
0-1      & 0.71 (0.022)      & 0.762 (0.011)  & SFTkatzAuto    \\\hline
0-2      & 0.699 (0.032)     & 0.739 (0.012)  & SFTkatzAuto    \\\hline
1-0      & 0.741 (0.013)       & 0.753 (0.012) & SFTkatz0.1-1.0 \\\hline
1-1      & 0.736 (0.034)      & 0.767 (0.007)  & SFTvanilla     \\\hline
1-2      & 0.754 (0.027)      & 0.767 (0.033)  & T              \\\hline
2-0      & 0.507 (0.045)      & 0.559 (0.009)  & SFTkatz0.0-0.5 \\\hline
2-1      & 0.599 (0.044)      & 0.618 (0.074)  & T              \\\hline
2-2      & 0.671 (0.017)      & 0.675 (0.014)  & SFTRPRauto     \\\hline
3-0      & 0.554 (0.018)      & 0.576 (0.009)  & SFTkatz0.0-0.5 \\\hline
3-1      & 0.517 (0.027)      & 0.553 (0.013)  & SFTGG0.0-1.0   \\\hline
3-2      & 0.515 (0.045)      & 0.555 (0.016)  & SFTRPRauto     \\\hline
4-0      & 0.49  (0.011)      & 0.508 (0.006)  & FTRPR0.0-0.5   \\\hline
4-1      & 0.51  (0.010)      & 0.532 (0.013)  & SFTkatz0.0-0.5 \\\hline
4-2      & 0.506 (0.010)      & 0.545 (0.009)  & FTkatz0.0-1.0  \\\hline
5-0      & 0.701 (0.028)      & 0.741 (0.005)  & FTkatz0.1-1.0  \\\hline
5-1      & 0.758 (0.010)      & 0.783 (0.005)  & SFTkatz0.1-1.0 \\\hline
5-2      & 0.756 (0.044)      & 0.807 (0.004)  & FTRPRauto      \\\hline
6-0      & 0.686 (0.009)      & 0.709 (0.010)  & FTkatz0.0-0.5  \\\hline
6-1      & 0.701 (0.030)      & 0.748 (0.018)  & SFTkatzAuto    \\\hline
6-2      & 0.726 (0.035)      & 0.759 (0.015)  & T              \\\hline
7-0      & 0.757 (0.010)      & 0.762 (0.008)  & SFTkatz0.0-0.5 \\\hline
7-1      & 0.738 (0.013)      & 0.764 (0.010)  & FTkatz0.0-0.5  \\\hline
7-2      & 0.731 (0.015)      & 0.761 (0.013)  & FTkatzAuto     \\\hline
8-0      & 0.677 (0.016)      & 0.724 (0.007)  & FTkatz0.0-0.5  \\\hline
8-1      & 0.717 (0.028)      & 0.752 (0.007)  & SFTkatz0.1-1.0 \\\hline
8-2      & 0.731 (0.016)      & 0.742 (0.013)  & SFTkatz0.0-0.5 \\\hline
9-0      & 0.699 (0.014)      & 0.747 (0.011)  & FTkatz0.0-1.0  \\\hline
9-1      & 0.68  (0.012)      & 0.726 (0.019)  & FTkatz0.0-0.5  \\\hline
9-2      & 0.627 (0.012)      & 0.726 (0.006)  & FTkatz0.1-1.0 \\ \hline
\end{tabular}
\smallskip
\caption{Accuracy (ACC) and standard deviation (SD) of the best vs original GCN model. Models written as: $F$- features only, $T$- topology only, $FT$- both features and topology, $FTvanilla$ - the original GCN. $S$ in the right column denotes usage of an additional feature weight matrix in the first layer (Equation~\ref{eq:gcn_feat_weight_mat}). The models that use $G_{NS}=\tilde{L}^{sym}_{NS}$, where $NS$ is a node-similarity ($Katz$, $RPR$, and $GG$) measure with different thresholds (Equation~\ref{katzLaplacian},~\ref{RPRLaplacian} and,~\ref{newton}) are represented in the last column with the corresponding node-similarity matrix (e.g. katz for the model FTkatz0.0-0.5). All the similarity-based $G$ are preprocessed and reconfigured based on Section~\ref{ak_preprocessing}. The preprocessing threshold $auto$ implies automatic selection of a threshold based on the mean value of the normalised node-similarity matrix (as per Section~\ref{ak_preprocessing}). Networks are represented in terms of snapshots, e.g. 0-0: first network's first snapshot, 0-1: first network's second snapshot etc.}
\label{tab:bestMethods}
\end{table}





\section{Conclusions and Future Work}

In this work, we have evaluated the performance of GCN on simulated friendship-based social network datasets. One limitation of the GCN is that it is limited to the number of neighbourhood path by the number of layers used in the model. We argued that using the node-similarity matrix as a graph representative allows us to solve this dependency between the $l^{th}$ layers and the $l^{th}$ order of the neighbourhood nodes. Additionally, our approach with the node-similarity measures may perform well enough with only a few layers compared with the original GCN due to the less dependency between the highest number of layers used in the model and the highest order of node neighbourhood considered. The GCN or any deep learning model is prone to overfitting when a large number of layers are used~\cite{kipf2016semi}, and our approach may get around this problem and achieve higher accuracy only with a few layers. It has also been empirically shown that most of the models with the augmented node-similarity measures outperform the original GCN.


In total we have proposed four new variations of the GCN model. Three of them are primarily based on the Katz, RPR, and GG scores as a form of the graph topology encoding. The fourth model is where we add learnable parameters for each of the features independent of the nodes for the entire graph, allowing the model to ignore the input features if it so chooses. This variation of the model can be used with the adjacency matrix as well as with the Katz, RPR or GG scores, and its primary motivation was the observation that for some datasets using the topology only, gives superior results. The results show that these new variations outperform the original GCN model in terms of accuracy.


For the node-similarity-based matrices, we have proposed a reconfiguration technique. This reconfiguration results in augmentation of the graph represented by the node-similarity matrix. This is particularly important as for node classification task with GCN-like models, we only have one graph sample to train the model. The augmentation technique can be used to better train the model on the same graphs with several different augmented node-similarity matrices (with different thresholds and similarity measurements). Several representations of the same graph topology can also work as a regularisation technique to prevent overfitting of the model. 

We argued that a node in Facebook-type social networks can be defined in terms of a set of preferences (which we coined as the \emph{sDNA}) of a node). Based on the \emph{sDNA}, our simulation strategy provides a comprehensive guideline on how to generate dynamic networks with features, and ground truth labels, particularly useful to train and test the neural network-based learning systems. We have validated the integration of features and topology of the simulated graphs based on the predictability of the GCN. If the integration process is good enough, the GCN should not perform better on a model with topology only, compared with a model with both the topology and feature. However, we have found that a large number of models would perform better with only the topology of the network. We have concluded that this is because not all the features play a similar role in the graph. To include such variation of importance for each of the features for all the nodes, we have introduced the weighted feature matrix for the GCN. The new variant of the GCN with weighted feature matrix, have shown to have great potential. With the weighted feature matrix, the majority (except in three datasets)  of the models perform better with both the feature and topology compared with the topology only. This not only produces a new variant of the GCN model but also shows that our integration process of the topology and features is successful.

The three cases where topology only performs better could be due to significantly more learnable parameters the model has compared with the feature and topology model that we have discussed. To solve this problem of an unequal number of learnable parameters, we have introduced another variation for the topology only model, where the number of learnable parameters is reduced by using a low-rank approximation of the weight matrix. The reduced parameter model for the topology only also performs well compared with the model with more parameters. A further inspection of those datasets may reveal the underlying reason why the topology only models perform well for them. However, it could be possible that for those three datasets, the features are reflected within the topology so well that the topology only model becomes more powerful and adding features simply results in redundancy.


We have used a few empirically selected thresholds for the augmented node-similarity matrix. A more effective way to select optimal thresholds is another future direction to explore.   

We also provide an opensource library for social network simulation written in Python with GPU computation support for high dimensional features. We aim to incorporate more features in the future for the simulation library. 

\section{References}

\bibliography{references}

\end{document}